\documentclass[sigconf]{acmart}

\copyrightyear{2025}
\acmYear{2025}
\setcopyright{cc}
\setcctype{by}
\acmConference[KDD '25]{Proceedings of the 31st ACM SIGKDD Conference on Knowledge Discovery and Data Mining V.2}{August 3--7, 2025}{Toronto, ON, Canada}
\acmBooktitle{Proceedings of the 31st ACM SIGKDD Conference on Knowledge Discovery and Data Mining V.2 (KDD '25), August 3--7, 2025, Toronto, ON, Canada}
\acmDOI{10.1145/3711896.3736892}
\acmISBN{979-8-4007-1454-2/2025/08}

\usepackage{algorithmic}
\usepackage[ruled,vlined,linesnumbered]{algorithm2e}
\usepackage{multicol}
\usepackage{multirow}
\usepackage{dsfont}
\usepackage{float}
\usepackage{booktabs}
\usepackage{makecell}
\usepackage{subfig}
\usepackage{graphicx}
\usepackage{caption}
\usepackage{hyperref}
\hypersetup{
    colorlinks=true,
    linkcolor=blue,
    citecolor=green,
    urlcolor=red
}
\usepackage{url}
\usepackage{xcolor}
\usepackage{bm}
\usepackage{enumitem}
\usepackage{etoc}
\etocdepthtag.toc{mtchapter}
\etocsettagdepth{mtchapter}{subsection}
\etocsettagdepth{mtappendix}{none}

\begin{document}
\title{Contrastive Graph Condensation: Advancing Data Versatility through Self-Supervised Learning}

\author{Xinyi Gao}
\affiliation{
\institution{The University of Queensland}
\city{Brisbane}
\country{Australia}
}
\email{xinyi.gao@uq.edu.au}

\author{Yayong Li}
\affiliation{
\institution{The University of Queensland}
\city{Brisbane}
\country{Australia}
}
\email{yayong.li@uq.edu.au}

\author{Tong Chen}
\affiliation{
\institution{The University of Queensland}
\city{Brisbane}
\country{Australia}
}
\email{tong.chen@uq.edu.au}

\author{Guanhua Ye}
\affiliation{
\institution{Beijing University of Posts and Telecommunications}
\city{Beijing}
\country{China}
}
\email{g.ye@bupt.edu.cn}

\author{Wentao Zhang}
\affiliation{
\institution{Peking University}  
\city{Beijing}
\country{China}
}
\email{wentao.zhang@pku.edu.cn}

\author{Hongzhi Yin}
\authornote{Corresponding author.}
\affiliation{
\institution{The University of Queensland}  
\city{Brisbane}
\country{Australia}
}
\email{h.yin1@uq.edu.au}

\renewcommand{\shortauthors}{Xinyi Gao et al.}

\begin{abstract}
With the increasing computation of training graph neural networks (GNNs) on large-scale graphs, graph condensation (GC) has emerged as a promising solution to synthesize compact, substitute graphs of the large-scale original graphs for efficient GNN training. However, these condensed graphs are specifically designed for the node classification task, significantly limiting the versatility of the synthesized data across various downstream tasks. This limitation predominantly stems from the reliance of existing GC methods on classification as the surrogate task for optimization, which leads to an excessive dependence on node labels and restricts their utility in label-scarcity scenarios. More critically, this surrogate task tends to overfit class-specific information within the condensed graph, consequently restricting the generalization capabilities of GC for other downstream tasks. To address these challenges, we introduce Contrastive Graph Condensation (CTGC), which adopts a self-supervised surrogate task to extract critical, causal information from the original graph and enhance the cross-task generalizability of the condensed graph. Specifically, CTGC employs a dual-branch framework to disentangle the generation of the node attributes and graph structures, where a dedicated structural branch is designed to explicitly encode geometric information through nodes' positional embeddings. By implementing an alternating optimization scheme with contrastive loss terms, CTGC promotes the mutual enhancement of both branches and facilitates high-quality graph generation through the model inversion technique. Extensive experiments demonstrate that CTGC excels in handling various downstream tasks with a limited number of labels, consistently outperforming state-of-the-art GC methods.
\end{abstract}


  
\maketitle

\section{Introduction}
Graph neural networks (GNNs)~\cite{gao2024accelerating,gao2023semantic} have been widely employed across complex systems to analyze graph-structured data, including chemical molecules \cite{guo2022graph}, social networks \cite{zhang2023comprehensive}, and recommender systems~\cite{long2023decentralized}. 
However, the exponential increase in graph data volume presents formidable challenges for training GNNs, particularly in scenarios that necessitate training multiple models, such as neural architecture search, continual learning, and federated learning. 
In addressing these challenges, graph condensation (GC) \cite{jin2022graph} has emerged as a promising approach. 
It synthesizes a compact yet representative condensed graph that serves as a substitute for the large-scale original graph during model training. By preserving the essential attributes of the original graph, GNNs trained on the condensed graph achieve performance comparable to those trained on the original graph while significantly reducing the training time and broadening applicability in resource-constrained scenarios \cite{yin2024device}.

\begin{table}[t]
\setlength{\abovecaptionskip}{1.5pt}
\renewcommand{\arraystretch}{1.1}
\centering
\caption{The performance of models trained on the condensed graph (GC)~\cite{jin2022graph}, the original graph with supervised learning (SL), and the original graph with our proposed self-supervised learning (SSL) task. The 3-shot setting is applied. Node classification (NC), link prediction (LP), and clustering (CL) are assessed using accuracy, AUC, and NMI, respectively.}
\resizebox{\linewidth}{!}
{
\begin{tabular}{l|ccc|ccc}
\Xhline{1.pt}
Dataset & \multicolumn{3}{c|}{Cora}      & \multicolumn{3}{c}{Citeseer}   \\ \hline
Task    & NC   & LP   & CL   & NC   & LP   & CL   \\ \hline
GC      & 55.6±3.0 & 84.3±0.7 & 35.2±2.8 & 55.9±0.8 & 87.4±0.3 & 31.5±1.7 \\
SL      & 65.9±4.3 & 87.4±1.5 & 38.6±2.0 & 57.2±2.4 & 88.6±1.0 & 32.4±0.8 \\
SSL     & 71.8±2.9 & 90.7±0.2 & 50.8±0.2 & 63.6±0.9 & 92.4±0.3 & 40.9±1.2 \\ \Xhline{1.pt}
\end{tabular}}
\label{prelimi}
\end{table}

Due to their efficacy in accelerating model training processes, GC methods have attracted substantial attention and achieved significant progress.
To synthesize condensed data, GC methods typically leverage a relay model to bridge the original and condensed graphs, deploying the \textit{classification} as the \textit{surrogate task} for optimization~\cite{jin2022condensing}.
Initially, GC methods utilize sophisticated optimization strategies and focus on matching parameters of the classification relay model, including model gradients~\cite{jin2022graph} and training trajectories~\cite{zheng_structure_free_2023} generated from both graphs. 
Subsequently, advanced studies expand these parameter matching methods to lighter label regression~\cite{wang2023fast} and class-prototype matching~\cite{zhao2023dataset}, further enhancing the efficiency and effectiveness of the condensation process.
Despite the diversity of existing GC approaches, the optimization processes in these methods consistently center on the classification surrogate task, and a recent study~\cite{gao2024rethinking} highlights that all these methods converge to the class distribution matching between the original and condensed graphs.
As a result, this uniform reliance on the classification surrogate task significantly limits the versatility of the synthesized data across other downstream tasks and constrains the real-world practicality of GC.

Specifically, the application of the classification surrogate task within GC optimization encounters two primary limitations.
On the one hand, the efficacy of GC is critically dependent on label availability, with existing methods premised on the assumption of abundant labels~\cite{liu2024gcondenser}.
Unfortunately, this assumption often contradicts real-world scenarios, where labels in large-scale graphs are costly to annotate and scarce.
This label scarcity leads to imprecise class representations, thereby diminishing the effectiveness of GC methods. 
On the other hand, the classification surrogate task in these methods tends to overfit class-specific information within the condensed graph, consequently restricting the capability of the condensed graph to support diverse downstream tasks~\cite{yang_does_2023}. 
As shown in Table \ref{prelimi}, models trained on condensed graphs under label sparsity (GC) consistently underperform those trained on the original graph (SL), not only in node classification but also across various downstream tasks.
In practice, downstream tasks vary significantly across real-world graph systems; for instance, recommendation systems model user-item interactions as link prediction tasks~\cite{yu2023self}, while clustering tasks are essential for analyzing item characteristics and user behaviors~\cite{chen2020try}. 
Therefore, it is imperative for GNNs trained on condensed graphs to demonstrate generalizability to a variety of downstream tasks within polytropic environments.
In a nutshell, the critical problem that arises for GC is: ``\textit{How can we effectively distill essential knowledge to the condensed graph without the dependency on class labels, and ensure the GNNs trained on condensed graphs maintain generalizability across diverse downstream tasks?}''

To tackle this problem, self-supervised learning (SSL)~\cite{10597920} provides a promising direction, as it inherently enables the learning of more transferable and adaptable representations without label availability, addressing the task-specific bias introduced by the classification surrogate task.
This effectiveness is illustrated in Table \ref{prelimi}, where the model trained on the original graph with the self-supervised task (SSL) outperforms the supervised learning model (SL) across the three tasks under the label sparsity issue. 
Despite the potential of SSL to enhance task generalization, developing a self-supervised method for GC remains an open area, which meets three critical objectives: 
(1) the surrogate task should extract the representative yet task-invariant information from the original graph in a self-supervised manner to circumvent task-specific biases introduced by class labels; (2) it should effectively summarize and compress the extracted information to facilitate the generation of a condensed graph; and (3) instead of using pairwise similarity between condensed node attributes~\cite{jin2022graph}, the condensed graph structure should be independently constructed~\cite{yang_does_2023} to topologically mimic the original graph, so as to offer stronger supplement predictive signals when training a label-free GNN for diverse tasks.
Although a recent attempt~\cite{wang2024self} tries to generate condensed graphs from any pre-trained GNN model, this method is designed for the graph-level datasets and fails to handle the node-level tasks. Moreover, the arbitrary selection of the graph-level pre-trained model cannot represent the ideal distribution for effective GC, potentially compromising the practical utility of the condensed graph.

In response to these objectives, we propose Contrastive Graph Condensation (CTGC), a self-supervised GC approach designed to efficiently handle diverse downstream tasks. As illustrated in Fig. \ref{figmain}, CTGC utilizes a dual-branch framework composed of semantic and structural branches, which are iteratively optimized through a unified contrastive surrogate task. Specifically, the semantic branch processes node attributes according to the graph structure to extract latent semantic information, while the structural branch explicitly encodes geometric information using the eigenvectors of the graph structure.
This disentanglement of information enables the independent construction of a condensed graph structure, thereby preserving the geometric characteristics of the original graph. Furthermore, the structural branch contributes additional information during the condensation process, compensating for the semantic information in the unsupervised scenario. For the sake of data condensation, two branches are optimized through clustering-based contrastive losses, which encourage the centroid embeddings to accurately represent the nodes within their respective clusters. Moreover, to promote mutual enhancement and alignment between branches, we propose an alternating optimization framework to optimize two branches iteratively with the exchange of cluster assignments.
Subsequently, the centroid embeddings from the two branches contain the comprehensive attributes of the original graph, and can be utilized to recover the node attributes and topological structures of the condensed graph through the model inversion technique~\cite{binici2024condensed}, respectively.
Consequently, CTGC eliminates the dependence on class labels in the condensation procedure and enables the independent generation of graph structures, thus facilitating high-quality condensed graphs and improving cross-task generalizability.

The main contributions of this paper are threefold:

\begin{itemize}[leftmargin=*]
\item \textbf{New observations and insights.} Recognizing that the classification surrogate task significantly restricts the versatility of graph condensation, we take the initiative to pioneer the investigation of the self-supervised graph condensation problem, successfully enhancing task generalization and mitigating label dependency.
\item \textbf{New methodology.} We present CTGC, the first self-supervised GC method designed to enhance the task generalizability of condensed graphs. CTGC introduces a novel dual-branch framework that disentangles semantic and structural information into distinct branches. Furthermore, our contrastive surrogate task facilitates the simultaneous extraction of transferable information and effective data compression.
\item \textbf{State-of-the-art performance.} Extensive experiments verify that CTGC excels in generating high-quality condensed graphs without label availability, surpassing various state-of-the-art GC methods across diverse downstream tasks. Our codes are available at: \href{https://github.com/XYGaoG/CTGCcode}{https://github.com/XYGaoG/CTGCcode}.
\end{itemize}

\section{Preliminaries}
In this section, we first revisit the fundamental concepts of GNNs, eigenvalue decomposition, and GC, and then formally define the problem studied.

\subsection{Graph Neural Networks}
\label{sec_gcn}
Consider that we have a large-scale graph $\mathcal{T}=\{{\bf A}, {\bf X}\}$ consisting of $N$ nodes. 
${\bf X}\in{\mathbb{R}^{N\times d}}$ denotes the $d$-dimensional node attribute matrix and ${\bf A}\in \mathbb{R}^{N\times N}$ is the adjacency matrix. 
We use ${\bf Y}\in{\mathbb{R}^{N\times C}}$ to denote the one-hot node labels over $C$ classes.
GNNs learn the embedding for each node by leveraging the graph structure and node attribute as the input.
Without loss of generality, we use graph convolutional network (GCN) \cite{DBLP:conf/iclr/KipfW17} as an example, where the convolution operation in the $k$-th layer is defined as follows: 
\begin{equation}
{\bf{H}}^{(k)} =\text{ReLU}\left(\hat{\mathbf{A}}{\bf H}^{(k-1)}{\mathbf W}^{(k)}\right),   
\label{eq_gcn}
\end{equation}
where ${\bf{H}}^{(k)}$ is the node embeddings of the $k$-th layer, and ${\bf{H}}^{(0)}=\mathbf{X}$.
$\hat{\mathbf{A}}=\widetilde{\mathbf{D}}^{-\frac{1}{2}}\widetilde{\mathbf{A}}\widetilde{\mathbf{D}}^{-\frac{1}{2}}$ is the normalized adjacency matrix. 
$\widetilde{\mathbf{A}}$ represents the adjacency matrix with the self-loop, $\widetilde{\mathbf{D}}$ denotes the degree matrix of $\widetilde{\mathbf{A}}$, and $\mathbf{W}^{(k)}$ is the trainable weights. 
$\text{ReLU}\left(\cdot\right)$ is the rectified linear unit function.
Afterward, the $K$-th layer embeddings ${\bf{H}}^{(K)}$ are predicted by specific prediction heads for different downstream tasks.

\subsection{Eigenvalue Decomposition}
Given the adjacency matrix ${\bf A}$, the normalized graph Laplacian is defined as ${\mathbf{L}}={\mathbf{I}}-{\mathbf{D}}^{-\frac{1}{2}}{\mathbf{A}}{\mathbf{D}}^{-\frac{1}{2}}$, where ${\mathbf{I}}$ is the identity matrix. The eigenvalue decomposition of graph Laplacian is defined as ${\mathbf{L}}={\bf U} {\bf{\Lambda}} {\bf U}^{\top}$, where ${\top}$ denotes the transpose operation. ${\bf{\Lambda}}$ is a diagonal matrix whose diagonal entries $0\leq\lambda_1\leq...\leq\lambda_N\leq2$ are the eigenvalues of ${\mathbf{L}}$. ${\bf U}=[ {\bf u}_1,...,{\bf u}_N] \in \mathbb{R}^{N\times N}$ are the corresponding eigenvectors.

The eigenvalues and eigenvectors encapsulate the geometric information and node positions within the graph topology, providing a comprehensive view of graph structural information. Specifically, the eigenvalues~\cite{lin2022spectrum} summarize key structural properties, such as connectivity~\cite{chung1997spectral}, clusterability~\cite{lee2014multiway}, and diffusion distance~\cite{hammond2013graph}. Meanwhile, the eigenvectors in ${\bf U}$ serve as \textbf{positional embeddings}~\cite{qiu2020gcc,zhu2024graphcontrol}, capturing the local structure associated with each node~\cite{von2007tutorial} in the graph.

\subsection{Graph Condensation}
\label{sec_gc}
Conventional graph condensation methods~\cite{jin2022graph,gao2025robgc,gao_graph_2023} are primarily developed within a supervised framework, aiming to generate a small condensed graph $\mathcal{S}=\{{\bf A'}, {\bf X'}\}$ with ${\bf A'}\in\mathbb{R}^{N'\times N'}$, ${\bf X'}\in\mathbb{R}^{N'\times d}$ as well as its label ${\bf Y'}\in{\mathbb{R}^{N'\times C}}$, where $N'\ll{N}$. The condensation ratio is denoted as $r=\frac{N'}{N}$.
GNNs trained on $\mathcal{S}$ can achieve comparable performance to those trained on the much larger $\mathcal{T}$. 
To connect the original graph $\mathcal{T}$ and condensed graph $\mathcal{S}$, a \textbf{relay model} $f_{\theta}$ parameterized by ${\theta}$ is employed in GC for encoding both graphs.
Concurrently, a \textbf{surrogate task} is introduced within the condensation process to facilitate the optimization of the relay model and the condensed graph.

Specifically, the classification task is predominantly employed as the surrogate task in existing GC methods, and the classification losses of $\mathcal{T}$ and $\mathcal{S}$ w.r.t. ${\theta}$ are defined as:
\begin{equation}
\label{eq_iniloss}
\begin{split}
&\mathcal{L} \left( \theta \right) = \ell \left( f_{\theta} \left( {\mathcal{T}} \right) , \mathbf{Y} \right),\\
&\mathcal{L}' \left( \theta \right)  = \ell \left( f_{\theta} \left( {\mathcal{S}} \right) , \mathbf{Y}' \right),  
\end{split}
\end{equation}
where $\ell\left( \cdot,\cdot \right)$ is the cross-entropy loss and $\mathbf{Y}'$ is predefined to match the class distribution in $\mathbf{Y}$.
Then the objective of GC can be formulated as a bi-level optimization problem:
\begin{equation}
\label{eq_oriloss}
\min_{\mathcal{S}} \mathcal{L}  \left( \theta^{\mathcal{S}} \right)
\:\: \text{s.t.}  \:\: \theta^{\mathcal{S}}   = \arg\min_{\mathcal{\theta}} \mathcal{L}'  \left( \theta \right) .
\end{equation}
The typical GC method~\cite{jin2022graph} introduces the gradient matching to solve this problem, i.e., aligning the model gradients at each training step generated from two graphs.  
Moreover, to simplify the optimization of the structure of condensed graph, it entangle the graph structure with the node attributes, parameterizing $\mathbf{A}'$ by $\mathbf{X}'$ as:
\begin{equation}
\label{eq:adj}
{\bf A}_{i,j}' = \sigma\left(\frac{{\text{MLP}([{\bf X}'_i; {\bf X}'_j])} + {\text{MLP}([{\bf X}'_j; {\bf X}'_i])}}{2}\right),
\end{equation}
where $\text{MLP}(\cdot)$ is a multi-layer perceptron (MLP), and fed with the concatenation of condensed node features ${\bf X}'_i$ and ${\bf X}'_j$. $\sigma$ denotes the sigmoid function.

In addition to gradient matching, various GC methods employ diverse optimization strategies to address the objective in Eq. (\ref{eq_oriloss}), including distribution matching~\cite{zhao2023dataset,xiao2025disentangled}, trajectory matching~\cite{zheng_structure_free_2023} and kernel ridge regression~\cite{wang2023fast,gao2024graphopen}.
While these methods demonstrate the potential to improve GC performance, they all deploy the classification as the surrogate task, which inherently impacts the cross-task generalizability of GC.

\subsection{Problem Formulation}
\label{sec_formulation}
To advance the condensed data versatility and mitigate the dependency on labels within the GC, we focus on a self-supervised GC framework targeting a large unlabeled graph $\mathcal{T} = \{\mathbf{A}, \mathbf{X}\}$. Our objective is to generate a small condensed graph $\mathcal{S}=\{{\bf A'}, {\bf X'}\}$ with $d$-dimensional target embeddings ${\bf H}'\in{\mathbb{R}^{N'\times d}}$, which serve as proxy labels of $\mathcal{S}$ to facilitate the pre-training of downstream GNNs. This condensed graph $\mathcal{S}$ expedites the model pre-training processes, allowing the pre-trained model to be fine-tuned efficiently for adaptation across various downstream tasks.

\begin{figure*}[t]
\setlength{\abovecaptionskip}{1pt}
\centering
\includegraphics[width=0.9\linewidth]{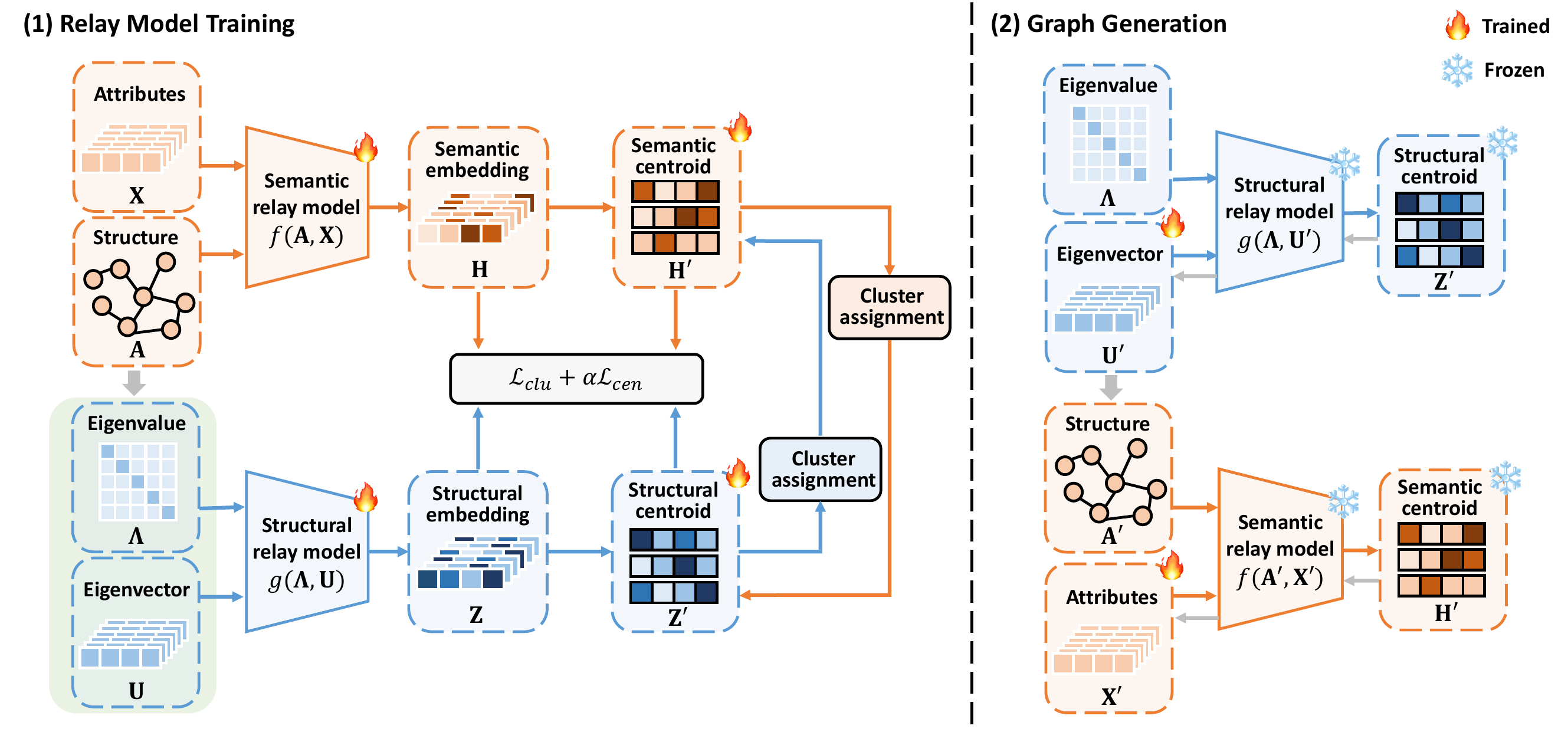}
\caption{The framework of our proposed CTGC, which comprises two stages: relay model training and graph generation. (1) CTGC employs a dual-branch architecture to separately extract semantic and structural information. The semantic relay model processes both the graph structure and node attributes, while the structural relay model uses eigenvalues and eigenvectors as inputs. These branches are iteratively optimized using contrastive losses. (2) The condensed graph is generated using the model inversion technique. This process begins with generating eigenvectors to construct the condensed graph structure, followed by learning node attributes based on the constructed graph structure.
}
\label{figmain}
\end{figure*}

\section{Methodologies}
We hereby present our proposed method, Contrastive Graph Condensation (CTGC), which comprises two stages: relay model training and graph generation, as illustrated in Fig. \ref{figmain}.
Initially, CTGC employs a dual-branch architecture to extract semantic and structural information separately.
Then we move on to the contrastive surrogate task to optimize the semantic and structural relay models, as well as to develop the centroid embeddings.
Finally, we generate the graph structure and node attributes using the relay models and centroid embeddings, culminating in the construction of the condensed graph.

\subsection{Dual-Branch Architecture}
The conventional condensation procedure described in Section \ref{sec_gc} employs a single relay model to encode both the graph structure information and the node attributes, with the structure of the condensed graph being generated based on condensed node attributes as Eq. (\ref{eq:adj}).
Although it simplifies the optimization of graph structure, the entanglement of condensed nodes and structure heavily caps the amount of preserved topological information from the original graph, as highlighted in~\cite{yang_does_2023,liugraph}. 
To resolve this issue and facilitate the independent extraction of structural information, we introduce an additional structural branch dedicated to encoding the graph structure. However, directly encoding the adjacency matrix of the large original graph is both computationally demanding and memory-intensive. To address this challenge, we construct the structural branch from the spectral perspective, ensuring the preservation of the geometric attributes in the original graph while enhancing computational efficiency.

Specifically, two distinct relay models are introduced to encode the original graph:
\begin{equation}
\label{encode}
\begin{split}
&{\bf{H}} = f({\bf A}, {\bf X}),\\
&{\bf{Z}} = g({\bf{\Lambda}}, {\bf U}),  
\end{split}
\end{equation}
where $f(\cdot)$ and $g(\cdot)$ represent the relay models for the semantic and structural branches, respectively. $f(\cdot)$ is the GCN model as used in conventional GC methods, while $g(\cdot)$ encodes the graph structure from the spectral perspective and processes the eigenvalues $\bf{\Lambda}$ and eigenvectors ${\bf U}$ of the original graph to produce the structural embeddings ${\bf Z} \in \mathbb{R}^{N \times d}$. To facilitate scalable encoding, we leverage EigenMLP~\cite{bo2024graph} as the structural relay model $g(\cdot)$ as follows: 
\begin{equation}
g({\bf{\Lambda}}, {\bf U})={\tilde{\bf{U}}} \rho ({\bf{\Lambda}}),
\end{equation}
where ${\tilde{\bf{U}}}$ is the sign-invariant eigenvectors and $\rho ({\bf{\Lambda}})$ represents the filtered eigenvalues. 
The sign-invariant eigenvectors ${\tilde{\bf{U}}}$ in EigenMLP are designed to mitigate the \textit{sign ambiguity}~\cite{bo2024graph} inherent in eigenvector decompositions, where arbitrary sign flips of eigenvectors can yield identical graph structures. Therefore, ${\tilde{\bf{U}}}$ takes both positive and negative forms of eigenvectors into consideration as:
\begin{equation}
{\tilde{\bf{U}}} = [\psi(\phi({\bf {u}}_i) + \phi({\bf {-u}}_i))]_{i=1}^{N},
\end{equation}
where $\psi(\cdot)$ and $\phi(\cdot)$ represent MLPs to transform the eigenvalues, and $[\cdot]$ denotes the concatenation operator.
$[ {\bf u}_1,...,{\bf u}_N] \in \mathbb{R}^{N\times N}$ are eigenvectors and ${\bf u}_i$ is the positional embedding for the node $i$.

Additionally, \textit{basis ambiguity}~\cite{bo2024graph} presents an another challenge in eigenvector decompositions, referring to the phenomenon where coordinate rotations of eigenvectors can lead to equivalent encoding outcomes for distinct graphs.
In responds to this issue, EigenMLP employs $\rho ({\bf{\Lambda}})$ to extend the eigenvalues ${\bf{\Lambda}}  = [\lambda_1, \lambda_2, ..., \lambda_N]$ into their high-dimensional Fourier features, thereby enhancing the model's ability to differentiate between graph structures:
\begin{equation}
\label{rho_loss}
\begin{split}
& \rho({\bf{\Lambda}}) = [\rho(\lambda_1), \rho(\lambda_2),..., \rho(\lambda_N)]^{\top}, \\
& \rho (\lambda_i) = [\sin(\lambda_i ), \cos(\lambda_i), ..., \sin(T \lambda_i), \cos(T \lambda_i)]{\bf W}_{\rho},
\end{split}
\end{equation}
where $T$ represents the period and ${\bf W}_{\rho} \in \mathbb{R}^{2T\times d}$ is a learnable weight matrix. $\lambda_i$ is the $i$-th eigenvalue in ${\bf{\Lambda}}$.

Although EigenMLP above effectively models the graph structure from a spectral perspective, employing full eigenvectors as the positional embeddings is impractical due to the inclusion of the dense and large matrix ${{\bf{U}}} \in \mathbb{R}^{N\times N}$, resulting in significant computational and memory costs for large graphs.
Among eigenvectors ${{\bf{U}}}$, those associated with the smallest and largest eigenvalues are critical for encapsulating geometric information. 
Specifically, eigenvectors corresponding to smaller eigenvalues emphasize the global community structure~\cite{mohar1991laplacian}. For example, the eigenvector corresponding to the second smallest eigenvalue demonstrates the connectivity of the graph and can be leveraged for graph partition by examining the signs or magnitudes of its entries, effectively capturing the global structure.
Conversely, the eigenvectors associated with larger eigenvalues represent higher-frequency components, highlighting finer structures and local variations within the graph \cite{nakatsukasa2013mysteries}.
These crucial eigenvectors play a pivotal role in graph modeling and are widely utilized in spectral GNNs~\cite{jin2020graph} and graph reduction methods~\cite{bo2023specformer,liugraph}.
To mitigate the computational burden of complete eigenvalue decomposition, we follow prior work~\cite{liugraph} by using the $K_1$ smallest and $K_2$ largest eigenvalues, along with their corresponding eigenvectors, as inputs to the structural branch. Moreover, the total number of eigenvalues is set to match the size of the condensed graph for subsequent graph generation, i.e., $K_1 + K_2 = N'$.
Specifically, the utilized eigenvalues and eigenvectors are denoted by ${\bf{\Lambda}}'  = [\lambda_1, ..., \lambda_{K_1},\lambda_{N-K_2},..., \lambda_{N}]$ and $ {\bf U'}=[ {\bf u}'_1,...,{\bf u}'_N] \in \mathbb{R}^{N\times N'}$, respectively. The original graph structure is encoded as:
\begin{equation}
g({\bf{\Lambda}}', {\bf U'})={\tilde{\bf{U}}'} \rho ({\bf{\Lambda}'}),
\end{equation} 
where ${\tilde{\bf{U}}}' = [\psi(\phi({\bf {u}'}_i) + \phi({\bf {-u}'}_i))]_{i=1}^{N}$. Given the significantly high condensation ratio in GC, where $N'\ll{N}$, the computational overhead is effectively kept under control.

\subsection{Contrastive Surrogate Task}

With the dual relay models, a surrogate task is crucial for extracting versatile information and enabling data compression. To achieve this, we design a clustering-based contrastive surrogate task that trains the relay models in a self-supervised manner while simultaneously performing compression within the latent space.
Without loss of generality, we illustrate the task using the semantic branch.

Specifically, given the semantic node embeddings ${\bf{H}}$, we initially utilize the K-means algorithm to group them into $N'$ clusters, i.e. ${\bf{\mathcal{C}}^{H}} = \{{\bf\mathcal{C}}^{\bf H}_1, {\bf\mathcal{C}}^{\bf H}_2, ..., {\bf\mathcal{C}}^{\bf H}_{N'}\}$, and use the average cluster embeddings as their respective centroids ${\bf{H'}} \in \mathbb{R}^{N'\times d}$, where ${\bf{H}}'_i=1/|{\bf\mathcal{C}}^{\bf H}_i|\sum_{j \in {\bf\mathcal{C}}^{\bf H}_i} {\bf{H}}_j$. 
Accordingly, each node $i$ is assigned to a cluster, with its cluster label defined as: ${{\bf{y}}^{\bf{H}}_i} =\arg\max_{1 \leq j \leq N'}{\text{sim}({\bf{H}}_i, {\bf{H}}'_j)}$.

This clustering process allows each node to be represented by its corresponding cluster centroid, whereby $N$ nodes in the original graph are condensed into $N'$ nodes in the latent space. However, clustering based on the node embeddings without training fails to yield optimal condensation results. Therefore, contrastive losses are designed to simultaneously optimize node distributions and centroid embeddings:
{\small
\begin{equation}
\begin{split}
& \mathcal{L}_{clu}({\bf{H}},{\bf{H}}',{{\bf{y}}^{\bf{H}}})=-\sum_{i=0}^{N}log\frac{\exp(\text{sim}({\bf{H}}_i,{\bf{H}}_{{\bf{y}}^{\bf{H}}_i}')/\tau)}{\sum_{j=0}^{N'} \mathds{1}_{[j \neq {\hat{\bf{y}}_{i}}]}\exp(\text{sim}({\bf{H}}_i, {\bf{H}}'_j)/\tau)},\\
& \mathcal{L}_{cen}({\bf{H}}')=-\sum_{i=0}^{N'}log\frac{\exp(\text{sim}({\bf{H}}'_i,{\bf{H}}'_{i})/\tau)}{\sum_{j=0}^{N'} \mathds{1}_{[j \neq i]}\exp(\text{sim}({\bf{H}}'_i, {\bf{H}}'_j)/\tau)},
\end{split}
\label{con_loss1}
\end{equation}}where $\text{sim}(\cdot, \cdot)$ is the cosine similarity and $\tau$ denotes the temperature. $\mathds{1}_{[j \neq i]} \in \{0, 1\}$ is an indicator function evaluating to 1 iff $j \neq i$. ${{\bf{y}}^{\bf{H}}_i}$ represents the cluster label of node $i$.
$\mathcal{L}_{clu}$ enhances cohesion within clusters by considering the centroid of node $i$ as the positive sample while treating other centroids as negative samples for node $i$. $\mathcal{L}_{cen}$ promotes separation among different centroids by discouraging similarities among different centroids.
Then, the relay model $f(\cdot)$ and centroid embeddings ${\bf{H'}}$ are optimized by the joint loss:
\begin{equation}
\label{loss}
\mathcal{L}({\bf{H}},{\bf{H}}',{{\bf{y}}^{\bf{H}}})= \mathcal{L}_{clu}({\bf{H}},{\bf{H}}',{{\bf{y}}^{\bf{H}}})+\alpha \mathcal{L}_{cen}({\bf{H}}'),
\end{equation}
where $\alpha$ is the weight to balance two losses. 
It is crucial to distinguish that, in contrast to typical contrastive losses in conventional self-supervised learning \cite{velivckovic2018deep}, our contrastive task excludes data augmentation and deliberately designed for data compression, thereby enhancing the representativeness of centroids and the efficacy of the condensation results.

Similarly, the structural relay model $g(\cdot)$ and centroid embeddings ${\bf{Z'}}$ can be optimized using the contrastive loss $\mathcal{L}({\bf{Z}},{\bf{Z}}',{{\bf{y}}^{\bf{Z}}})$.

\noindent\textbf{Branch Alignment.} 
The dual-branch architecture learns node representations with diverse emphasis, conditioned on semantic features and structural similarities, respectively. Therefore, the nodes with similar semantic (or structural) information will be clustered together and represented by the corresponding cluster centroid.
This may lead to inconsistencies in cluster assignments, where a node's positioning relative to the generated clusters may differ. Consequently, this discrepancy can impede the alignment of condensed node attributes with the graph structure, ultimately resulting in a corrupted condensed graph.

To address this issue, we introduce an alternating optimization framework to align the two branches, as detailed in Algorithm \ref{al}. 
During the training procedure, cluster labels are exchanged between branches, and each branch is optimized to learn the cluster assignments inferred by the other.
In the semantic branch, we fix $g(\cdot)$ and update $f(\cdot)$ with the cluster labels ${{\bf{y}}^{\bf{Z}}}$ inferred by $g(\cdot)$, allowing the structural information learned by $g(\cdot)$ to be distilled into $f(\cdot)$: 
\begin{equation}
\label{eq:semloss}
\mathcal{L}_{sem}= \mathcal{L}({\bf{H}},{\bf{H}}',{{\bf{y}}^{\bf{Z}}}).
\end{equation}
In the structural branch, we fix $f(\cdot)$, and $g(\cdot)$ is optimized using the cluster labels ${{\bf{y}}^{\bf{H}}}$ inferred by $f(\cdot)$:
\begin{equation}
\label{eq:strloss}
\mathcal{L}_{str}= \mathcal{L}({\bf{Z}},{\bf{Z}}',{{\bf{y}}^{\bf{H}}}).
\end{equation}
The two branches are trained iteratively, allowing for their progressive alignment.
As a central component of the training process, the cluster labels, ${{\bf{y}}^{\bf{H}}}$ and ${{\bf{y}}^{\bf{Z}}}$, not only capture node correlations w.r.t semantic and structural features but also represent the condensation relationships between cluster centroids and individual nodes.  
The semantic branch’s labels, ${{\bf{y}}^{\bf{H}}}$ , enable the structural branch to learn semantic similarities among nodes, while the structural branch’s labels,  ${{\bf{y}}^{\bf{Z}}}$ , guide the semantic branch in capturing structural similarities. As a result, both semantic and structural information are effectively distilled across branches. This process ensures the consistency of optimization results and enhances condensation representations. The empirical validation of this alternating optimization method is presented in Section \ref{alter_sec}.

\subsection{Graph Generation}

The contrastive surrogate task effectively encodes the semantic and structural information into centroid embeddings.
This self-supervised task acts as a condensation mechanism, where each centroid embedding (i.e., ${\bf{H'}}$ and ${\bf{Z'}}$) aggregates the collective features of all node embeddings within its cluster.
Consequently, with the alignment of branches, the graph structure of the condensed graph can be recovered from ${\bf{Z'}}$ due to the independent graph structure encoding of $g(\cdot)$, and node attributes can be constructed by referring to ${\bf{H'}}$.

Specifically, we utilize the model inverse technique~\cite{binici2024condensed}, which optimizes the input data according to the fixed model and target outputs. As shown in Fig. \ref{figmain} (b), we first generate the eigenvectors ${\bf U'}$ of the condensed graph according to the well-trained relay model $g(\cdot)$, target centroid embedding ${\bf{Z'}}$ and the utilized eigenvalues of the original graph ${\bf{\Lambda}}'$:
\begin{equation}
\label{generate1}
\arg\min_{{\bf U'}}||{\bf{Z'}}- g({\bf{\Lambda}}', {\bf U'})||_2+||{\bf{I}}- {\bf U'}^{\top}{\bf U'}||_2,
\end{equation}
where the second term ensures the natural orthogonality of the generated eigenvectors. Afterwards, the condensed graph is recovered according to ${\bf U'}$ and ${\bf{\Lambda}}'$:
\begin{equation}
\label{adj_cond}
{\bf{A'}}= {\bf{I}}-{\bf U'} {\bf{\Lambda}'} {\bf U'}^{\top}.
\end{equation}
The attributes for the condensed graph ${\bf X'}$ is derived according to the relay model $f(\cdot)$, centroid embedding ${\bf{H'}}$ and condensed graph structure ${\bf{A'}}$:
\begin{equation}
\label{x_cond}
\arg\min_{{\bf X'}}||{\bf{H'}}- f({\bf A'}, {\bf X'}) ||_2.
\end{equation}
Finally, the condensed graph $\mathcal{S}=\{{\bf A'}, {\bf X'}\}$ is employed in training downstream GNN models, with ${\bf{H'}}$ serving as the training labels.

\begin{algorithm}[t]
\SetAlgoVlined
\small
\textbf{Input:} Original graph $\mathcal{T}=\{{\bf A}, {\bf X}\}$.\\
\textbf{Output:} Semantic relay model $f(\cdot)$, structural relay model $g(\cdot)$, centroid embeddings ${\bf H}'$ and ${\bf Z}'$, condensed graph $\mathcal{S}=\{{\bf A'}, {\bf X'}\}$.\\

Calculate ${\bf{\Lambda}}$ and $ {\bf U}$, then encode the graph by Eq. (\ref{encode}).\\ 
Initialize $f(\cdot)$ and calculate ${\bf H}'$ by Section \ref{initial}.\\
Initialize ${{\bf{y}}^{\bf{Z}}} \leftarrow {{\bf{y}}^{\bf{H}}}$.\\
\For{$iter=1,\ldots,K_{iter}$}  
{
\For{$m=1,\ldots,M_{train}$}  
{
Compute $\mathcal{L}_{sem}=\mathcal{L}({\bf{H}},{\bf{H}}',{{\bf{y}}^{\bf{Z}}})$ with Eq. (\ref{eq:semloss}).\\
Update $f(\cdot)$ and ${\bf H'}$.\\
}
Compute ${{\bf{y}}^{\bf{H}}_i}=\arg\max_{1 \leq j \leq N'}{\text{sim}({\bf{H}}_i, {\bf{H}}'_j)}$.\\
\For{$m=1,\ldots,M_{train}$}{
Compute $\mathcal{L}_{str}=\mathcal{L}({\bf{Z}},{\bf{Z}}',{{\bf{y}}^{\bf{H}}})$ with Eq. (\ref{eq:strloss}).\\
Update $g(\cdot)$ and ${\bf Z'}$.\\
}
Compute ${{\bf{y}}^{\bf{Z}}_i}=\arg\max_{{1 \leq j \leq N'}}{\text{sim}({\bf{Z}}_i, {\bf{Z}}'_j)}$.\\
}
Generate condensed graph $\mathcal{S}$ by Eq. (\ref{generate1})-(\ref{x_cond}).
\caption{Process of CTGC}
\label{al}
\end{algorithm}

\subsection{Further Detailed Analysis}
\label{initial}
\noindent\textbf{Initialization.} 
Despite the contrastive surrogate task enabling the summaries of extracted information, the initial cluster labeling impacts the subsequent optimization results.
Hence, to enhance the convergence, we follow previous work~\cite{zheng2022rethinking, liu2023dink} to pre-train the semantic relay model by a simple SSL task and generate the initial cluster labels.
Specifically, $\bf{X}$ is shuffled by randomly altering the node order, and the modified attributes $\bf{\tilde{X}}$ are then encoded through the incorrect graph structure to obtain augmented embeddings ${\bf{\tilde{H}}} =  f({\bf A}, \bf{\tilde{X}})$.
These augmented embeddings are employed to construct a binary classification task alongside ${\bf{{H}}}$, with the relay model $f(\cdot)$ being pre-trained to distinguish between them. This process enhances the discriminative capacity of the latent space, thereby providing an effective initialization.
After training, initial cluster labels are derived by applying K-means to ${\bf{H}}$, and the semantic relay model is utilized for subsequent optimizations.

\noindent\textbf{Algorithm.} 
The detailed process of CTGC is shown in Algorithm \ref{al}. Initially, we perform eigenvalue decomposition and encode the original graph (line 3). Then, the semantic branch is initialized, and ${\bf{{H}'}}$ is calculated using K-means. In line 5, we initialize ${{\bf{y}}^{\bf{Z}}}$ with ${{\bf{y}}^{\bf{H}}}$ to prepare for the initial contrastive training of the semantic branch. Subsequently, the semantic branch (lines 7-10) and the structural branch (lines 11-14) are iteratively optimized. Finally, the condensed graph is generated according to both branches.

\noindent\textbf{Time Complexity.} The time complexity of CTGC comprises three main components: training of the semantic branch, structural branch, and computation of graph generalization. 
Firstly, the time complexity for training the semantic relay model in the semantic branch is $\mathcal{O}(L|E|d+Nd)$, where $L$ denotes the number of layers, $|E|$ is the number of edges, and $d$ represents the dimensionality of embeddings.
For the structural branch, the time complexity for training the structural relay model is $\mathcal{O}(NTd)$, where $T$ is the period of polynomial in EigenMLP.
Additionally, the complexity of decomposing the $N'$ smallest or largest eigenvalues of the original graph is $\mathcal{O}(N'N^2)$.
The clustering step incurs a complexity of $\mathcal{O}(N'Ndt)$, with $t$ denoting the number of iterations for K-means. 
Lastly, the complexity for graph generalization is $\mathcal{O}(N'Td+LN'^2d)$.

Therefore, the clustering and eigenvalue decomposition constitute the primary computational burdens in CTGC.
However, the scalability of CTGC is facilitated by the reduced size of the condensed graph and the integration of well-established acceleration libraries, such as FAISS~\cite{douze2024faiss} and Scipy~\cite{gommers2022scipy}.

\begin{table*}[t]
\setlength{\abovecaptionskip}{1pt}
\renewcommand{\arraystretch}{1.1} 
\centering
\caption{The performance of graph reduction methods on 3-shot node classification (NC), link prediction (LP) and clustering (CL) tasks. $r$ is the condensation ratio. The Whole Dataset for NC and CL indicates GNNs are trained on the original graph using few-shot labels. While Whole Dataset for LP refers to GNNs trained from \textit{scratch} with prediction heads on the original graph.}
\label{all_performance}
\resizebox{\linewidth}{!}
{
\begin{tabular}{l|c|cccccccc|c|c}
\Xhline{1.pt}
\multirow{2}{*}{Dataset ($r$)}       & \multicolumn{1}{l|}{\multirow{2}{*}{Task}} & \multicolumn{8}{c|}{$\{{\bf A}, {\bf X}, {\bf Y}\}$}    & $\{{\bf A}, {\bf X}\}$ & \multirow{2}{*}{\begin{tabular}[c]{@{}l@{}}Whole\\ Dataset\end{tabular}} \\ \cline{3-11}
     & \multicolumn{1}{l|}{}      & Kcenter  & Herding  & Coarsening & GCond    & GCDM     & SimGC    & GCSR     & GDEM     & Ours   &  \\ \hline
\multirow{3}{*}{\begin{tabular}[c]{@{}l@{}}Cora \\ (2.6\%)\end{tabular}}     & NC & 50.3±2.5 & 54.8±0.9 & 52.7±4.4   & 55.6±3.0 & 53.3±1.2 & 57.6±1.5 & 61.0±2.3 & 60.8±1.9 & \textbf{70.0±2.5}      & 65.9±4.3 \\
     & LP & 84.5±1.2 & 83.5±1.0 & 83.5±1.5   & 84.3±0.7 & 85.6±0.7 & 84.2±1.0 & 85.7±0.6 & 84.2±1.0 & \textbf{90.8±0.4}      & 87.4±1.5 \\
     & CL & 31.0±4.5 & 35.3±2.8 & 38.5±2.2   & 35.2±2.8 & 34.4±2.4 & 34.0±1.0 & 36.1±2.0 & 35.9±1.2 & \textbf{48.8±1.9}      & 38.6±2.0 \\ \hline
\multirow{3}{*}{\begin{tabular}[c]{@{}l@{}}Citeseer \\ (1.8\%)\end{tabular}} & NC & 49.0±5.2 & 52.7±2.6 & 54.7±4.6   & 55.9±0.8 & 55.8±1.4 & 55.4±1.2 & 56.3±1.7 & 55.6±1.7 & \textbf{63.2±1.8}      & 57.2±2.4 \\
     & LP & 85.1±0.3 & 86.4±1.1 & 87.5±2.0   & 87.4±0.3 & 87.2±2.7 & 87.0±1.0 & 88.4±0.1 & 87.5±1.0 & \textbf{92.3±0.5}      & 88.6±1.0 \\
     & CL & 25.2±2.8 & 30.6±4.8 & 29.6±2.0   & 31.5±1.7 & 32.9±1.5 & 33.4±1.6 & 33.5±2.3 & 34.0±1.5 & \textbf{40.4±1.7}      & 32.4±0.8 \\ \hline
\multirow{3}{*}{\begin{tabular}[c]{@{}l@{}}Arxiv \\ (0.25\%)\end{tabular}}   & NC & 39.7±5.7 & 42.8±2.9 & 38.1±1.6   & 44.6±0.4 & 44.6±3.5 & 44.8±1.5 & 45.8±1.1 & 45.2±2.1 & \textbf{47.8±0.7}      & 46.3±3.7 \\
     & LP & 92.3±0.7 & 94.8±0.3 & 89.9±1.8   & 92.2±1.2 & 93.0±2.5 & 94.2±1.6 & 93.0±0.7 & 93.2±1.6 & \textbf{95.1±0.4}      & 96.3±0.1 \\
     & CL & 34.5±0.7 & 36.3±0.7 & 33.9±0.0   & 34.3±0.1 & 35.2±0.5 & 33.0±0.5 & 34.1±0.9 & 34.9±0.5 & \textbf{37.2±0.3}      & 35.1±1.1 \\ \hline
\multirow{3}{*}{\begin{tabular}[c]{@{}l@{}}Reddit \\ (0.1\%)\end{tabular}}   & NC & 68.3±3.4 & 74.0±2.9 & 59.1±4.0   & 70.0±1.2 & 70.9±1.4 & 70.2±1.8 & 72.7±1.2 & 73.1±0.6 & \textbf{86.4+0.2}      & 82.5±0.8 \\
     & LP & 76.0±0.6 & 76.5±0.9 & 77.1±0.8   & 84.0±2.3 & 85.9±1.2 & 85.4±1.4 & 87.0±2.1 & 86.9±1.3 & \textbf{94.6+0.1}      & 95.0±0.3 \\
     & CL & 60.5±0.4 & 60.9±1.0 & 55.7±0.9   & 59.0±2.0 & 58.1±1.3 & 60.0±1.5 & 59.7±1.7 & 59.9±1.4 & \textbf{70.7+0.9}      & 63.2±0.8 \\ \hline
\multirow{3}{*}{\begin{tabular}[c]{@{}l@{}}Products\\ (0.025\%)\end{tabular}} & NC        & 26.4±1.7 & 28.4±1.0 & 25.9±1.3 & 29.2±1.1 & 29.0±1.3 & 31.9±1.2 & 32.4±1.1 & 32.2±1.6 & \textbf{34.2±0.7}  & 32.8±0.6                       \\
                          & LP        & 93.9±1.0 & 94.7±1.1 & 93.0±1.2 & 94.4±1.4 & 94.8±1.0 & 95.7±1.1 & 94.9±1.0 & 95.5±1.3 & \textbf{96.5±0.8}  & 96.7±0.1                       \\
                          & CL        & 29.2±1.2 & 33.2±0.8 & 28.8±1.0 & 32.9±0.8 & 32.7±0.7 & 33.7±0.8 & 33.7±1.2 & 33.4±1.1 & \textbf{38.3±0.9}  & 33.8±0.6   \\
     
     \Xhline{1.pt}
\end{tabular}
}
\end{table*}

\section{Experiments}
We design comprehensive experiments to validate the effectiveness of our proposed CTGC and aim to answer the following questions. 

\noindent\textbf{Q1}: Compared to other graph reduction methods, can CTGC achieve better performance across different downstream tasks?
\textbf{Q2}: Can the alternating optimization method align two branches and enhance the performance?
\textbf{Q3}: Can the condensed graph generalize well to different GNN architectures? 
\textbf{Q4}: How do the different components, i.e., structural branch, initialization, alternative optimization, constraints, and graph construct method affect CTGC?
\textbf{Q5}: How does the semantic relay model generated by CTGC perform?
\textbf{Q6}: What are the characteristics of the condensed graph?
\textbf{Q7}: How does CTGC perform under different condensation ratios?
\textbf{Q8}: How do the different hyper-parameters affect CTGC?

Due to space limitations, we delay \textbf{Q6}-\textbf{Q8} to \textbf{Appendix \ref{app_condratio}-\ref{app_statistic_sec}}.

\subsection{Experimental Settings}
\label{exp_set}

\noindent\textbf{Datasets \& Baselines}. 
We assess our proposed methods using 5 datasets: Cora~\cite{DBLP:conf/iclr/KipfW17}, Citeseer~\cite{DBLP:conf/iclr/KipfW17}, Ogbn-arxiv (Arxiv)~\cite{hu2020open}, Reddit~\cite{DBLP:conf/iclr/ZengZSKP20} and Ogbn-products (Products)~\cite{hu2020open}.
Our proposed method is compared with 8 graph reduction methods, including coreset (Herding~\cite{welling2009herding} and K-Center~\cite{sener2017active,farahani2009facility}), coarsening \cite{loukas2019graph,huang2021scaling}, gradient matching-based GC (GCond~\cite{jin2022graph}), distribution matching-based GC (GCDM~\cite{liu2022graph} and SimGC~\cite{xiao2024simple}), trajectory matching-based GC (GCSR~\cite{liu2024graph}), and eigenbasis matching-based GC (GDEM~\cite{liugraph}).

Notably, the compared baselines necessitate the node labels during the graph generation process, whereas our method condenses the graph in an unsupervised manner. We utilize latent GNN embeddings for node selection in coreset methods. Importantly, due to the significant dependency of coreset and coarsening methods on the node labels, we initially train a GNN using few-shot labels and subsequently expand the label set by the pseudo labels. For GC baselines, the graph is condensed using few-shot labels. 

\noindent\textbf{Evaluation Protocol}. To evaluate the cross-task generalizability of models trained on condensed graphs, we utilize three commonly employed downstream tasks: node classification (NC), link prediction (LP), and node clustering (CL). The performances of these tasks are measured by accuracy, AUC and NMI score, respectively. Following graph self-supervised learning paradigms~\cite{velivckovic2018deep}, we freeze the model parameters trained on the condensed graph, thereby converting it into a static feature extractor. Subsequently, dedicated prediction heads for node classification and link prediction are trained using these features. For clustering, the K-means algorithm is directly applied to the node embeddings.

To facilitate rapid adaptation of prediction heads, we assess the node classification and link prediction tasks under the few-shot scenario, which is consistent with the objective of efficient model training in GC. 
Specifically, the node classification task is evaluated under 3-shot and 5-shot settings \footnote{This differs from the graph few-shot learning \cite{ding2020graph}, and we do not engage in the meta learning paradigm.}, wherein 3 and 5 labels are provided per class for training the prediction head, respectively. Unless otherwise specified, the 3-shot setting is evaluated by default.
Notably, there is \textit{no validation set} in CTGC, and the optimal model is determined based on the lowest training loss.
For link prediction task, we follow~\cite{sun2023all} and provide 100 links as the training set for prediction head training. $5\%$ are used for validation and $15\%$ for testing. All remaining links are solely used for message-passing. 

It is crucial to underscore that all validation and test edges are excluded from the original graph used in the condensation process to prevent information leakage.

For reproducibility, all hyper-parameters are summarized in \textbf{Appendix \ref{sec_hyperpara}}.

\subsection{Evaluation Across Multiple Tasks (\textbf{Q1})}

\noindent\textbf{Node Classification Task.}
We assess the node classification capability of CTGC against baseline methods in handling the label sparsity issue. We present both 3-shot and 5-shot evaluation settings, and their results are shown in Table \ref{all_performance} and Table \ref{nc5} in \textbf{Appendix \ref{app_5shot}}, respectively.
In these tables, the Whole Dataset (WD) indicates that GNNs are trained on the original graph using few-shot labels, which suffers from substantial computational costs due to the large scale of the original graph.
Our proposed CTGC method consistently outperforms other baselines by a large margin, notably also surpassing WD at substantial compression rates. For instance, given the 3-shot setting, CTGC achieves improvements of 1.5\% over WD at the condensation rate of 0.25\% on the Arxiv dataset. This performance advantage extends across all other datasets, with the largest improvement observed on Citeseer, where CTGC achieves 63.2\%, compared to 57.2\% for WD. Similar trends are observed in the 5-shot setting, as shown in Table \ref{nc5} in in {Appendix \ref{app_5shot}}.
In terms of other baselines, their performance varies across different settings, with all of them falling short of the results achieved by WD and our method. Although Herding exhibits the best performance among the coreset methods, its best performance only reaches 54.8\% on Cora, significantly lower than our result of 70.0\%. 
These findings highlight the efficacy of our self-supervised GC paradigm, which preserves critical information in the condensed graph for model pre-training, reducing the dependency on label quantity.

\begin{figure}[t]
\setlength{\abovecaptionskip}{1pt}
\centering
\includegraphics[width=0.85\linewidth]{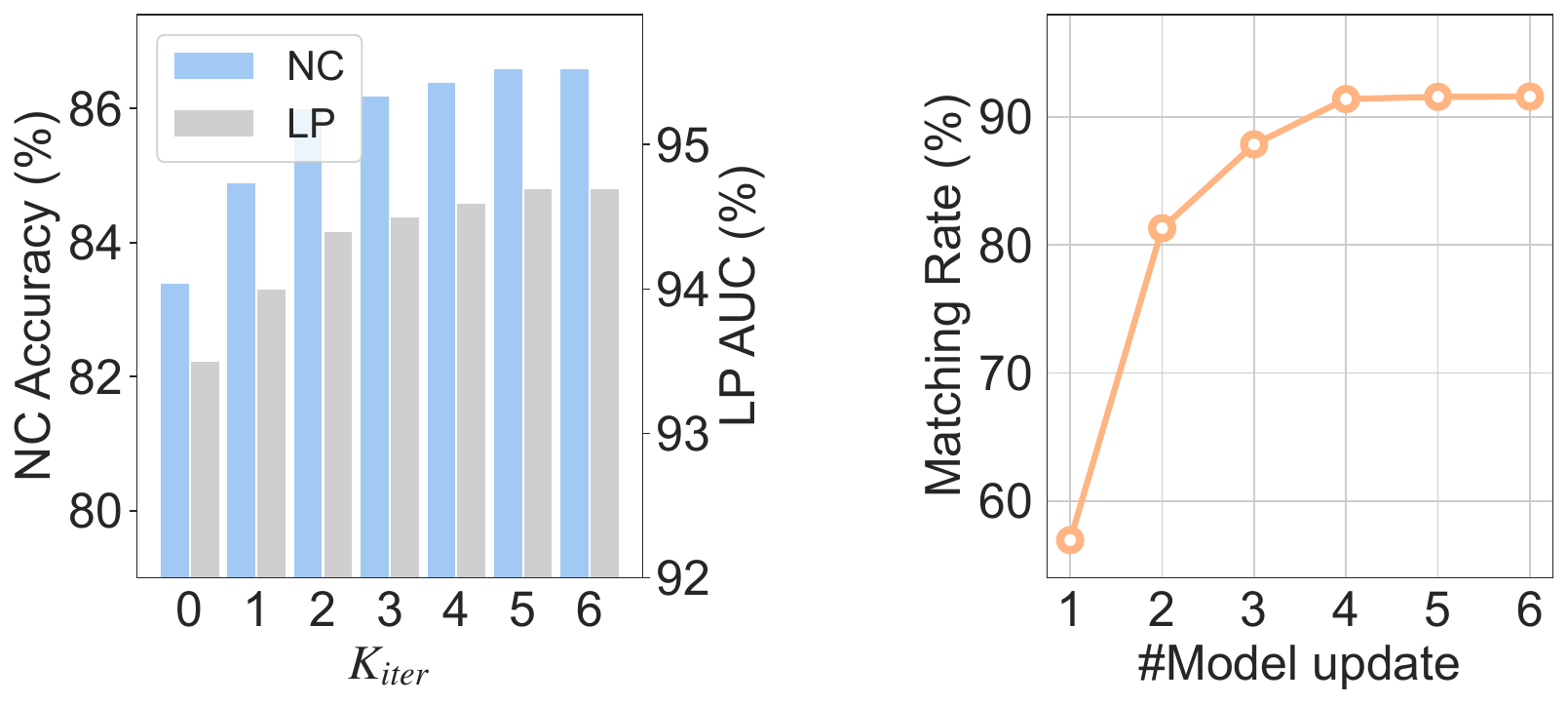}
\caption{The effect of the alternating optimization on the Reddit ($r$=0.1\%). Left: task performances. Right: accuracy of assignment alignment.}
\label{fig_iter}
\end{figure}

\begin{table}[t]
\setlength{\abovecaptionskip}{1pt}
\renewcommand{\arraystretch}{1.1}
\centering
\caption{The generalizability for GNN architectures on the Arxiv ($r$=0.25\%). AVG represents the average value.}
\label{cross_arxiv}
\resizebox{0.95\linewidth}{!}
{
\begin{tabular}{l|l|rrrrr|r}
\Xhline{1.pt}
Task& Method & \multicolumn{1}{l}{GCN} & \multicolumn{1}{l}{SGC} & \multicolumn{1}{l}{SAGE} & \multicolumn{1}{l}{APPNP} & \multicolumn{1}{l|}{Cheby} & \multicolumn{1}{l}{AVG} \\ \hline
\multirow{6}{*}{NC} & GCond  & 44.6                    & 42.5                     & 44.6                    & 43.3                      & 38.1& 42.6                    \\
                    & GCDM   & 44.6                    & 42.8                     & 42.8                    & 42.9                      & 39.6& 42.5                    \\
                    & SimGC  & 44.8                    & 44.1                     & 43.0                    & 42.0                      & 40.6& 42.9                    \\
                    & GCSR   & 45.8                    & 45.4                     & 44.9                    & 43.6                      & 40.9& 44.1                    \\
                    & GDEM   & 45.2                    & 45.7                     & 45.1                    & 43.8                      & 41.1& 44.2                    \\
                    & Ours   & \textbf{47.8}           & \textbf{47.8}            & \textbf{46.3}           & \textbf{46.5}             & \textbf{41.8}              & \textbf{46.0}           \\ \hline
\multirow{6}{*}{LP} & GCond  & 92.2                    & 89.0                     & 94.4                    & 93.4                      & 91.5& 92.1                    \\
                    & GCDM   & 93.0                    & 93.7                     & 84.4                    & 92.5                      & 92.3& 91.2                    \\
                    & SimGC  & 94.2                    & 91.6                     & 92.8                    & 93.2                      & 92.9& 92.9                    \\
                    & GCSR   & 93.0                    & 93.5                     & 92.5                    & 93.3                      & \textbf{94.1}              & 93.3                    \\
                    & GDEM   & 93.2                    & 93.3                     & 92.5                    & 92.3                      & 93.5& 93.0                    \\
                    & Ours   & \textbf{95.1}           & \textbf{94.6}            & \textbf{94.7}           & \textbf{94.9}             & {93.2}              & \textbf{94.5}           \\ \hline
\multirow{6}{*}{CL} & GCond  & 34.3                    & 35.9                     & 35.1                    & 32.7                      & 25.1& 32.6                    \\
                    & GCDM   & 35.2                    & 35.7                     & 35.6                    & 34.3                      & 28.4& 33.8                    \\
                    & SimGC  & 33.0                    & 33.3                     & 32.1                    & 30.7                      & 28.1& 31.4                    \\
                    & GCSR   & 34.1                    & 33.8                     & 35.3                    & 32.8                      & 29.7& 33.1                    \\
                    & GDEM   & 34.9                    & 34.2                     & 35.0                    & 32.1                      & 30.0& 33.2                    \\
                    & Ours   & \textbf{37.2}           & \textbf{37.2}            & \textbf{35.9}           & \textbf{35.3}             & \textbf{31.2}              & \textbf{35.4}    \\ \Xhline{1.pt}
\end{tabular}}
\end{table}

\noindent\textbf{Link Prediction Task.}
The link prediction model is established by appending a trainable prediction head on top of the frozen pre-trained model derived from reduced graphs under the 3-shot setting. 
In contrast, WD provides a benchmark for comparison, which differs from other tasks in training the GNN model with prediction heads \textit{from scratch} on the original graph, rather than merely training the prediction heads on top of a pre-trained GNN model. 
Compared to the baselines, our method demonstrates significant improvements on all four datasets. Notably, it outperforms WD on the Cora and Citeseer datasets, achieving improvements of 3.4\% and 3.7\%. On the Arxiv and Reddit datasets, our method yields comparable results to WD, while substantially surpassing other baselines. These results suggest that existing GC methods are highly dependent on node labels and fail to effectively preserve link correlations in the condensed graph. In contrast, by leveraging self-supervised learning, our proposed method generates a more robust relay model that is devoid of node classification bias, resulting in superior link prediction performance.

\begin{table}[t]
\setlength{\abovecaptionskip}{1pt}
\renewcommand{\arraystretch}{1.1}
\centering
\caption{The ablation study of CTGC on model modules, constraints, and graph generation method.}
\label{ablation}
\resizebox{\linewidth}{!}
{
\begin{tabular}{l|ccc|ccc}
\Xhline{1.pt}
Dataset& \multicolumn{3}{c|}{Arxiv ($r$=0.25\%)} & \multicolumn{3}{c}{Reddit ($r$=0.1\%)} \\ \hline
Task  & NC                & LP                & CL                & NC                & LP                & CL                \\ \hline
Ours  & \textbf{47.8±0.7} & \textbf{95.1±0.4} & \textbf{37.2±0.3} & \textbf{86.4±0.2} & \textbf{94.6±0.1} & \textbf{70.7±0.9} \\ \hline
Ours w KNN                   & 46.6±0.7  & 94.9±0.9 & 35.7±0.3 & 84.8±0.4 & 93.5±0.6 & 67.9±1.1 \\
Ours w/o $\mathcal{L}_{dis}$ & 46.8±1.1  & 94.8±0.4 & 37.0±0.1 & 85.7±0.2 & 94.2±0.1 & 69.4±1.0 \\
Ours w/o INIT                & 46.3±0.3  & 95.0±0.6 & 36.5±0.3 & 86.2±0.4 & 94.4±0.4 & 70.3±1.1 \\
Ours w/o ITER                & 45.9±0.7  & 94.9±0.3 & 37.0±0.4 & 83.4±0.1 & 94.3±0.8 & 67.1±0.5 \\
Ours w/o STR                & 45.6±1.0  & 94.5±0.6 & 36.4±0.3 & 82.9±0.4 & 94.2±0.6 & 66.9±0.3 \\ \Xhline{1.pt}
\end{tabular}}
\end{table}

\noindent\textbf{Clustering Task.}
Beyond the supervised tasks, we also evaluate our model on the node clustering task. This unsupervised task is undertaken based on the pre-trained model derived from the 3-shot setting without additional training. We remain the consistent setting with all other baselines for a fair comparison. 
Overall, our method achieves superior performance over all other baselines, including WD, by a significant margin. The greatest improvement is observed on the Cora dataset, where our method achieves an 10.2\% improvement compared to the best baseline (WD). This validates the effectiveness of our method on the clustering task.

\subsection{Alternating Optimization Method (\textbf{Q2})}
\label{alter_sec}
In this section, we examine the impact of alternating optimization method of CTGC across various downstream tasks. We change the value of $K_{iter}$, which dictates the number of iterations for the alternating optimization of the semantic and structural branches. Fig. \ref{fig_iter} (left) shows the node classification and link prediction performance on the Reddit dataset. The increment of $K_{iter}$ generally enhances model performance and leads to gradual convergence. However, higher values of $K_{iter}$ lead to greater computational demands, necessitating the selection of an appropriate value to balance performance gains with computational complexity.

To evaluate the convergence of the optimization process, Fig. \ref{fig_iter} (right) shows the matching rate of cluster labels between the two branches. The matching rate gradually increases and eventually converges as the model updates, reflecting the alignment of the two branches and the stability of the optimization process.

\subsection{Generalizability for GNN Architectures (\textbf{Q3})}
 
To evaluate the generalizability of our proposed method across different GNN architectures, we train various models on the condensed graph, including GCN, SAGE~\cite{hamilton2017inductive}, SGC~\cite{pmlr-v97-wu19e}, APPNP~\cite{bojchevski2020pprgo}, and Cheby~\cite{defferrard2016convolutional}. The performances for these GC methods on the Arxiv are presented in Table \ref{cross_arxiv}. The results indicate that all tested GNN models are effectively trained using the condensed graphs and achieve comparable performance across different tasks. Specifically, GCN and SGC show superior performance as these models utilize the same convolution kernel as the relay model during the condensation process. Our method consistently delivers the best results across different architectures and downstream tasks, demonstrating the robustness and versatility of our generated condensed graphs.

\begin{table}[t]
\setlength{\abovecaptionskip}{1pt}
\renewcommand{\arraystretch}{1.1}
\centering
\caption{The performance comparison of the semantic relay model across various tasks under the 3-shot setting. Node classification (NC), link prediction (LP), and clustering (CL) are assessed using accuracy, AUC, and NMI, respectively.}
\label{teacher}
\resizebox{0.8\linewidth}{!}
{
\begin{tabular}{l|c|ccc}
\Xhline{1.pt}
         Dataset ($r$)            &  Task  & \begin{tabular}[c]{@{}l@{}}Whole \\ Dataset\end{tabular} & \begin{tabular}[c]{@{}l@{}}Semantic \\relay model\end{tabular} & Ours \\  \hline
\multirow{3}{*}{\begin{tabular}[c]{@{}l@{}}Cora\\ (2.60\%)\end{tabular}}       & NC   & 65.9±4.3   & 71.8±2.9          & 70.0±2.5 \\
          & LP   & 87.4±1.5   & 90.7±0.2          & 90.8±0.4 \\
          & CL   & 38.6±2.0   & 50.8±0.2          & 48.8±1.9 \\ \hline
\multirow{3}{*}{\begin{tabular}[c]{@{}l@{}}Citeseer\\ (1.80\%)\end{tabular}}   & NC   & 57.2±2.4   & 63.6±0.9          & 63.2±1.8 \\
          & LP   & 88.6±1.0   & 92.4±0.3          & 92.3±0.5 \\
          & CL   & 32.4±0.8   & 40.9±1.2          & 40.4±1.7 \\ \hline
\multirow{3}{*}{\begin{tabular}[c]{@{}l@{}}Arxiv\\ (0.25\%)\end{tabular}} & NC   & 46.3±3.7   & 47.9±0.4          & 47.8±0.7 \\
          & LP   & 96.3±0.1   & 96.4±0.6          & 95.1±0.4 \\
          & CL   & 35.1±1.1   & 37.4±0.2          & 37.2±0.3 \\ \hline
\multirow{3}{*}{\begin{tabular}[c]{@{}l@{}}Reddit\\ (0.10\%)\end{tabular}}     & NC   & 82.5±0.8   & 86.6±0.1          & 86.4±0.2 \\
          & LP   & 95.0±0.3   & 94.7±0.4          & 94.6±0.1 \\
          & CL   & 63.2±0.8   & 70.8±0.4          & 70.7±0.9 \\ \Xhline{1.pt}
\end{tabular}}
\end{table}

\subsection{Ablation Study (\textbf{Q4})}

To validate the impact of individual components, CTGC is evaluated by disabling specific components. Specifically, we evaluate three types of components and detailed results are shown in Table \ref{ablation}.

\noindent\textbf{Graph Generation Method.}
Contrary to traditional GC methods that construct the condensed graph structure based on node attributes, CTGC utilizes positional embeddings, which are independent of the node attributes of the condensed graph. To compare these graph structure generation methods, we evaluate CTGC by replacing the condensed graph generation with the KNN graph~\cite{zhu2021deep} and denote this method as ``w KNN''. It is constructed by directly measuring the similarity of condensed node attributes. In contrast, our proposed method leverages diverse eigenvalues and eigenvectors in graph construction, preserving the spectral properties of the original graph and containing stronger generalizability.

\noindent\textbf{Constraint.}
CTGC is evaluated by disabling the centroid discrimination loss $\mathcal{L}_{dis}$ (``w/o $\mathcal{L}_{dis}$''). According to Table \ref{ablation}, we can observe that $\mathcal{L}_{dis}$ improves cluster distribution and node classification performance, confirming that the centroid discrimination loss facilitates the uniform distribution of clustering centroids.

\noindent\textbf{Method Modules.}
CTGC is assessed in the following configurations: (1) ``w/o INIT'': without relay model initialization; (2) ``w/o ITER'': without iterated optimization of the semantic and structural branches. (3) ``w/o STR'': without the structural branch and generate the graph structure by KNN graph. 
The removal of all these method modules leads to noticeable declines in different tasks, underscoring the necessity in condensation procedure.
The initialization lays a robust foundation for effective clustering and optimization of the relay model. Iterated optimization enables the structural correlations among positional embeddings to be transferred to the semantic branch, thereby enhancing the relay model and node representations. Notably, the most substantial performance decline occurs when the structural branch is removed. For instance, the node classification and clustering performances decline 3.5\% and 3.8\% on Reddit dataset, underscoring the necessity of incorporating eigenvectors to explicitly encode structural information.

\subsection{Relay Model Performance (\textbf{Q5})}
\label{relay_per}

We also examine the performance of the semantic relay model across various downstream tasks. 
As illustrated in Table \ref{teacher}, the self-supervised learning paradigm significantly enhances model generalization. Notably, on datasets such as Cora, Citeseer, and Arxiv, the link prediction performance of the relay model surpasses that of models trained from scratch. 
Furthermore, our clustering-based surrogate task facilitates the effective compression of extracted information during the training procedure of relay models. This is pivotal in generating high-quality condensed graphs, and the performance discrepancy between the relay model and the models trained on the condensed graph is well controlled.

\subsection{Visualization (\textbf{Q6})}
\label{relay_per}

\begin{figure}[t]
\setlength{\abovecaptionskip}{1pt}
\centering
\begin{minipage}[t]{0.42\linewidth}
    \centering
    \includegraphics[width=\linewidth]{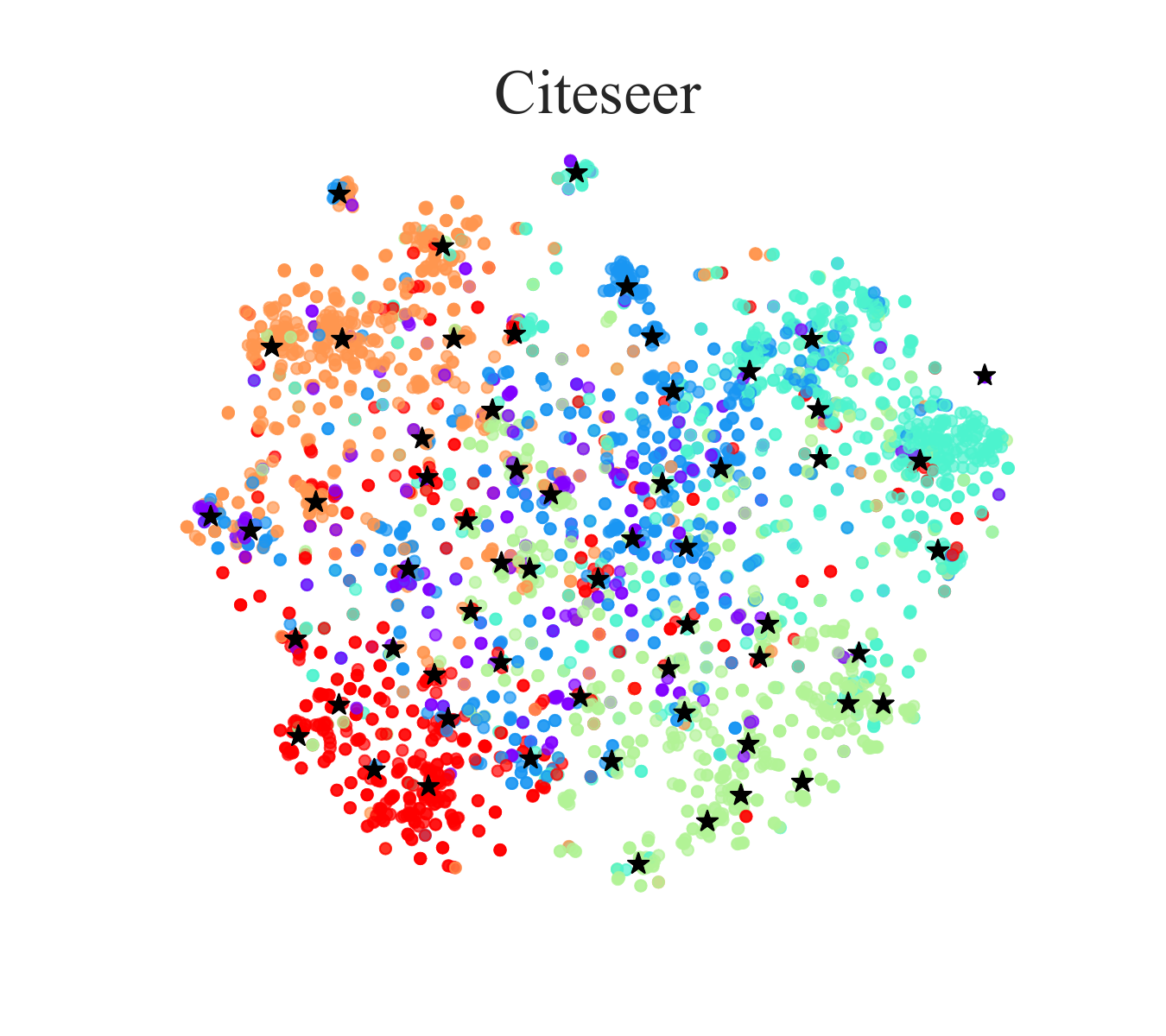}
\end{minipage}
\begin{minipage}[t]{0.42\linewidth}
    \centering
    \includegraphics[width=\linewidth]{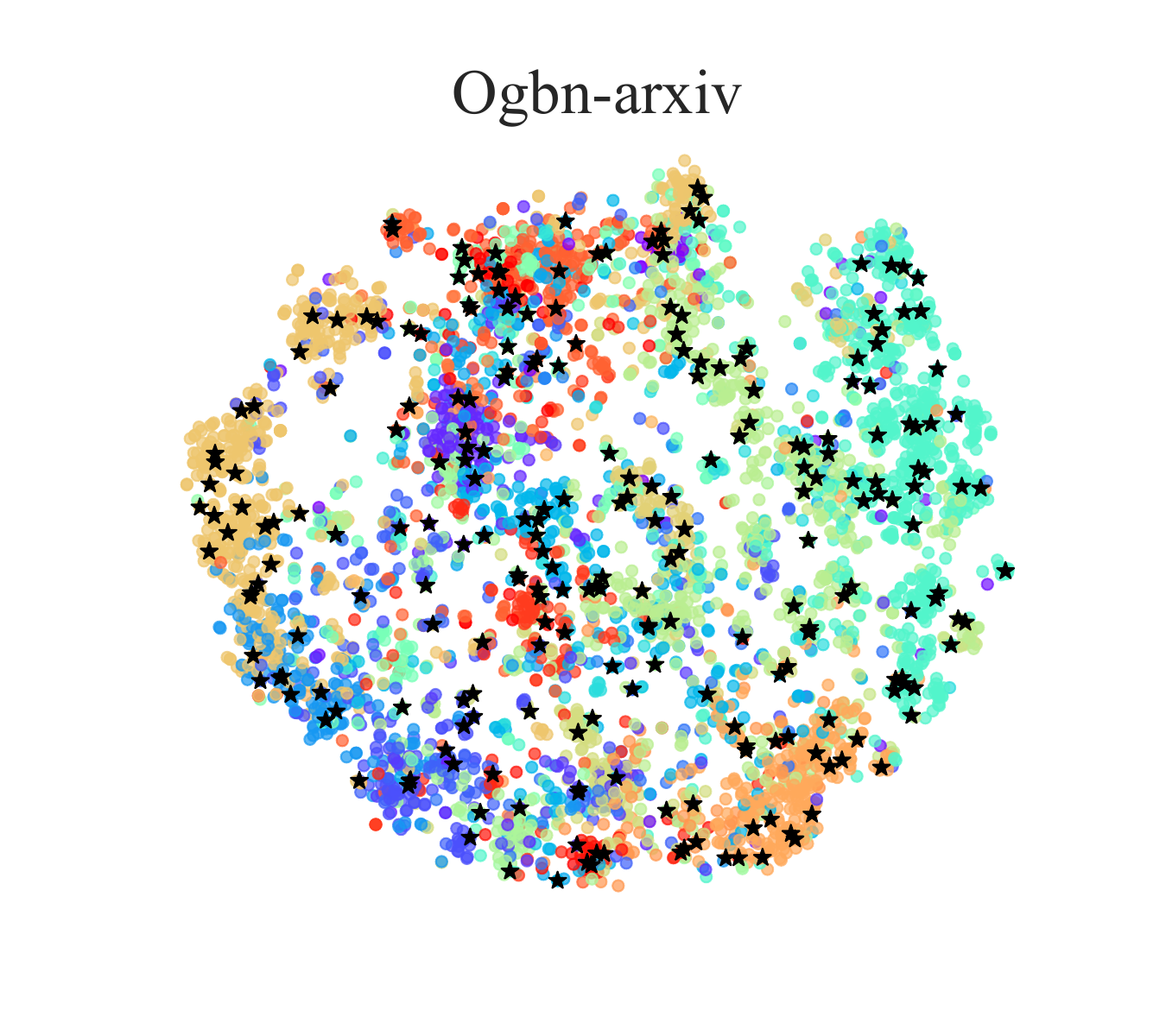}
\end{minipage}
\caption{Node distribution of original and condensed graphs. Condensed nodes (black stars) are label-free, while original nodes are color-coded by class labels.}
\label{fig:hyper_beta}
\end{figure}

To compare the distribution of the original and condensed graphs, we visualize the nodes in the two graphs in Figure \ref{fig:hyper_beta}. The condensed nodes are uniformly distributed and align well with the community structure of the original graph.

\section{Conclusion}

In this paper, we present Contrastive Graph Condensation (CTGC), a novel self-supervised GC approach to efficiently handle diverse downstream tasks. CTGC employs a dual-branch framework with a unified self-supervised surrogate task to separately extract latent semantic and geometric information. Moreover, the condensed graph is generated using the model inversion technique, which eliminates the dependence on class labels in the GC process. The future research could focus on developing foundation GC methods applicable across various datasets.
\section*{Acknowledgment}
The Australian Research Council partially supports this work under the streams of Future Fellowship (Grant No. FT210100624), Discovery Early Career Researcher Award (Grants No. DE230101033), the Discovery Project (Grant No. DP240101108 and DP240101814), and the Linkage Projects (Grant No. LP230200892 and LP240200546).

\bibliographystyle{ACM-Reference-Format}
\bibliography{ref}

\appendix
\section{Appendix}

\subsection{Experimental Settings} 
\label{sec_hyperpara}
Following the existing GC works, we employ a two-layer GCN as the semantic relay model, and for evaluation, two-layer GNNs with 256 hidden units are utilized. Unless otherwise specified, GCN is evaluated by default.
The hyper-parameter configurations for our proposed method are detailed in Table \ref{hypertable}. 
Here, $M_{pre}$ and $M_{train}$ denote the number of epochs allocated for pre-training and training of the relay model, respectively.
$K_{iter}$ represents the number of training iterations. The learning rates for pre-training, semantic relay model training, and structural relay model training are denoted by $lr_{pre}$, $lr_{sem}$, and $lr_{str}$, respectively. $\alpha$ is used to balance the losses in Eq. (\ref{loss}). $\tau$ denotes the temperature in the contrastive loss. $K_1$ and $K_2$ are defined as $0.9N'$ and $0.1N'$ for all datasets, respectively.

\subsection{5-Shot Node Classification Task (\textbf{Q1})}
\label{app_5shot}

In addition to the 3-shot classification setting, we further evaluate our method under a 5-shot node classification task, and the results are presented in Table \ref{nc5}. With more training labels, the classification performance of the Whole Dataset (WD) and various graph reduction methods consistently improves compared to the 3-shot setting. Notably, our proposed method achieves a more significant performance gain on the Arxiv dataset, with a 1.6\% improvement under a 0.25\% condensation ratio. These results demonstrate the superiority of the model trained on our condensed graph.

\begin{table}[t]
\setlength{\abovecaptionskip}{1pt}
\renewcommand{\arraystretch}{1.35}
\centering
\caption{The hyper-parameter configurations for evaluated datasets.}
\label{hypertable}
\resizebox{\linewidth}{!}
{
\begin{tabular}{lrrrrrrrrr}
\Xhline{1.pt}
           & \multicolumn{1}{l}{$M_{pre}$} & \multicolumn{1}{l}{$lr_{pre}$} & \multicolumn{1}{l}{$M_{train}$} & \multicolumn{1}{l}{$K_{iter}$} & \multicolumn{1}{l}{$lr_{sem}$} & \multicolumn{1}{l}{$lr_{str}$} & \multicolumn{1}{l}{$\alpha$}  & \multicolumn{1}{l}{$\tau$} \\ \hline
Cora       & 200    & 0.001   & 20       & 5    & 0.0001  & 0.001                    &1000             & 0.3 \\
Citeseer   & 200    & 0.001   & 20       & 3        & 0.0001  & 0.001    &1000                    & 0.3 \\
Arxiv      & 200    & 0.001   & 50       & 3         & 0.0001  & 0.1           &1000          & 0.3 \\
Reddit     & 20     & 0.0001  & 40       & 3        & 0.001   & 0.1    &10000                & 0.3 \\ 
Products   & 200     & 0.001   & 10       & 2         & 0.0001  & 0.001           &1000         & 0.3 \\ 
\Xhline{1.pt}
\end{tabular}}
\end{table}

\begin{figure}[t]
\setlength{\abovecaptionskip}{1pt}
\centering
\includegraphics[width=0.9\linewidth]{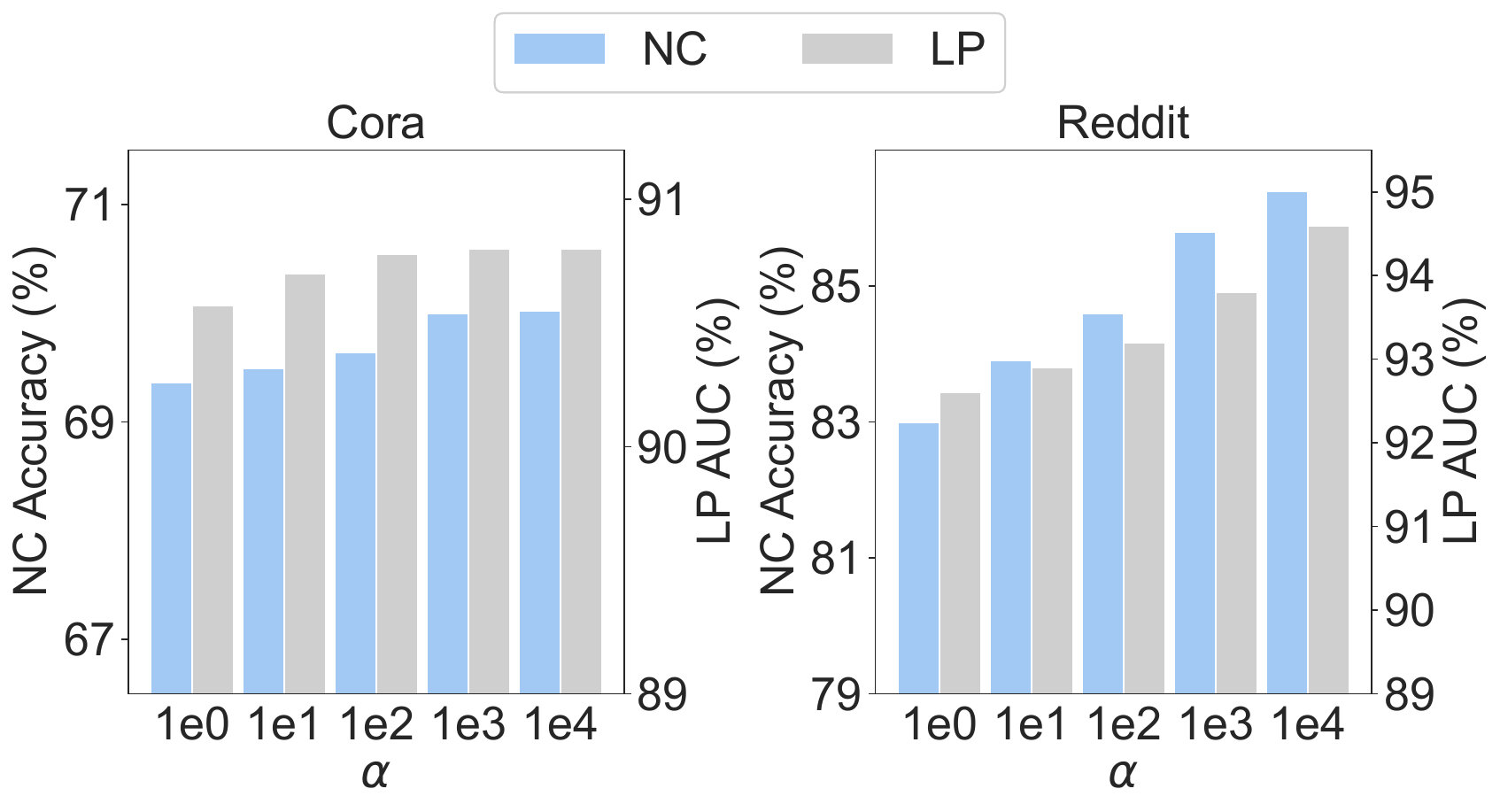}
\caption{The effect of hyper-parameter $\alpha$ on the Cora ($r$=2.6\%) and Reddit ($r$=0.1\%) under the 3-shot setting.}
\label{fig:hyper}
\end{figure}

\subsection{Different Condensation Ratios (\textbf{Q7})}
\label{app_condratio}
To evaluate the generalizability of our method to condensation ratios, we evaluate three different condensation ratios for Reddit as utilized in GCond \cite{jin2022graph}. The Table \ref{nc3} demonstrates the results for 3-shot node classification, link prediction, and clustering tasks. Specifically, Herding is sensitive to the condensation ratio, with a larger number of nodes in the reduced graph generally yielding better outcomes. In contrast, our proposed method and GC baselines exhibit robustness to the condensation ratio across different tasks, and maintain superior performance even with smaller condensed graph sizes. These findings align with conventional GC methods \cite{gao2024graph}, further highlighting the superiority of GC over other graph reduction techniques.

\begin{table*}[t]
\setlength{\abovecaptionskip}{1pt}
\renewcommand{\arraystretch}{1.1}
\centering
\caption{The 5-shot node classification accuracy (\%) of different graph reduction methods. The Whole Dataset indicates GNNs are trained on the original graph using few-shot labels.}
\label{nc5}
\resizebox{\linewidth}{!}
{
\begin{tabular}{l|c|cccccccc|c|c}
\Xhline{1.pt}
\multirow{2}{*}{Dataset} & \multicolumn{1}{l|}{\multirow{2}{*}{$r$}} & \multicolumn{8}{c|}{$\{{\bf A}, {\bf X}, {\bf Y}\}$}                                    & \multicolumn{1}{l|}{$\{{\bf A}, {\bf X}\}$} & \multicolumn{1}{l}{\multirow{2}{*}{\begin{tabular}[c]{@{}l@{}}Whole\\ Dataset\end{tabular}}} \\ \cline{3-11}
                         & \multicolumn{1}{l|}{}                     & Kcenter  & Herding  & Coarsening & GCond    & GCDM     & SimGC    & GCSR     & GDEM     & Ours                                        & \multicolumn{1}{l}{}                                                                         \\ \hline
Cora                     & 2.60\%                                    & 49.8±2.1 & 61.6±2.5 & 58.6±1.2   & 58.5±2.2 & 59.4±0.7 & 58.6±1.5 & 63.4±0.7 & 61.3±2.6 & \textbf{71.1±1.1}                           & 66.4±3.2                                                                                     \\
Citeseer                 & 1.80\%                                    & 48.8±2.1 & 52.9±1.5 & 58.0±4.4   & 54.1±2.2 & 55.8±2.0 & 55.8±1.4 & 56.5±1.8 & 56.6±1.8 & \textbf{63.5±0.9}                           & 60.8±3.8                                                                                     \\
Arxiv                    & 0.25\%                                    & 39.6±1.4 & 42.2±2.4 & 38.2±0.4   & 44.7±1.2 & 45.8±2.4 & 44.8±1.9 & 45.3±1.6 & 45.2±1.3 & \textbf{48.5±0.8}                           & 46.9±3.7                                                                                     \\
Reddit                   & 0.10\%                                    & 71.7±4.7 & 75.0±3.0 & 60.8±1.8   & 70.6±2.6 & 70.6±3.2 & 71.2±1.8 & 72.9±1.9 & 73.8±1.3 & \textbf{87.2+0.2}                           & 84.2±0.9                                                                                     \\ \Xhline{1.pt}
\end{tabular}
}
\end{table*}

\begin{table*}[t]
\setlength{\abovecaptionskip}{1pt}
\renewcommand{\arraystretch}{1.1} 
\centering
\caption{The condensation performances under different condensation ratios $r$ on Reddit. 3-shot setting is employed.}
\label{nc3}
\resizebox{\linewidth}{!}
{
\begin{tabular}{l|c|cccccccc|c|c}
\Xhline{1.pt}
\multirow{2}{*}{Task} & \multicolumn{1}{l|}{\multirow{2}{*}{$r$}} & \multicolumn{8}{c|}{$\{{\bf A}, {\bf X}, {\bf Y}\}$}                                    & \multicolumn{1}{l|}{$\{{\bf A}, {\bf X}\}$} & \multicolumn{1}{l}{\multirow{2}{*}{\begin{tabular}[c]{@{}l@{}}Whole\\ Dataset\end{tabular}}} \\ \cline{3-11}
                      & \multicolumn{1}{l|}{}                     & Kcenter  & Herding  & Coarsening & GCond    & GCDM     & SimGC    & GCSR     & GDEM     & Ours                                        & \multicolumn{1}{l}{}                                                                         \\ \hline
\multirow{3}{*}{NC}   & 0.05\%                                    & 69.0±5.9 & 65.7±1.6 & 48.6±3.0   & 70.3±1.8 & 70.0±2.0 & 71.7±1.1 & 71.8±1.1 & 72.4±0.4 & \textbf{85.2+0.5}                           & \multirow{3}{*}{82.5±0.8}                                                                    \\
                      & 0.10\%                                    & 68.3±3.4 & 74.0±2.9 & 59.1±4.0   & 70.0±1.2 & 70.9±1.4 & 70.2±1.8 & 72.7±1.2 & 73.1±0.6 & \textbf{86.4+0.2}                           &                                                                                              \\
                      & 0.20\%                                    & 68.8±3.4 & 74.2±2.2 & 65.9±1.5   & 70.4±0.4 & 70.7±1.0 & 69.8±1.0 & 71.6±1.1 & 73.1±0.3 & \textbf{86.6+0.7}                           &                                                                                              \\ \hline
\multirow{3}{*}{LP}   & 0.05\%                                    & 75.2±1.3 & 75.9±0.9 & 75.0±3.5   & 85.6±2.0 & 86.3±1.3 & 85.8±2.7 & 86.1±2.9 & 86.5±1.7 & \textbf{93.9+0.4}                           & \multirow{3}{*}{95.0±0.3}                                                                    \\
                      & 0.10\%                                    & 76.0±0.6 & 76.5±0.9 & 77.1±0.8   & 84.0±2.3 & 85.9±1.2 & 85.4±1.4 & 87.0±2.1 & 86.9±1.3 & \textbf{94.6+0.1}                           &                                                                                              \\
                      & 0.20\%                                    & 77.4±0.6 & 79.2±0.5 & 78.1±1.9   & 83.6±2.3 & 87.4±1.4 & 86.4±2.3 & 86.1±0.7 & 87.5±1.2 & \textbf{94.2+0.1}                           &                                                                                              \\ \hline
\multirow{3}{*}{CL}   & 0.05\%                                    & 59.6±1.5 & 59.2±0.4 & 43.2±2.0   & 59.9±1.3 & 58.1±1.2 & 59.3±1.4 & 58.7±0.3 & 59.7±1.2 & \textbf{67.6+0.5}                           & \multirow{3}{*}{63.2±0.8}                                                                    \\
                      & 0.10\%                                    & 60.5±0.4 & 60.9±1.0 & 55.7±0.9   & 59.0±2.0 & 58.1±1.3 & 60.0±1.5 & 59.7±1.7 & 59.9±1.4 & \textbf{70.7+0.9}                           &                                                                                              \\
                      & 0.20\%                                    & 60.4±1.5 & 61.5±1.2 & 60.1±1.2   & 62.4±1.6 & 61.1±1.0 & 62.7±0.2 & 61.3±1.1 & 62.3±1.6 & \textbf{70.1+0.5}                           &  \\                                                                                          \Xhline{1.pt}
\end{tabular}
}
\end{table*}

\begin{table*}[t]
\setlength{\abovecaptionskip}{1pt}
\renewcommand{\arraystretch}{1.1}
\centering
\caption{The task performance and statistical comparison between the original graph and our condensed graph. The 3-shot setting is applied.}
\label{stat}
\resizebox{0.7\linewidth}{!}
{
\begin{tabular}{l|rr|rr|rr|rr}
\Xhline{1.pt}
Dataset & \multicolumn{2}{c|}{Cora ($r$=2.6\%)} & \multicolumn{2}{c|}{Citeseer ($r$=1.8\%)}& \multicolumn{2}{c|}{Arxiv ($r$=0.25\%)} & \multicolumn{2}{c}{Reddit ($r$=0.1\%)} \\ \hline
Graph & Original & Ours& Original & Ours& Original& Ours& Original & Ours \\ \hline
\#Nodes  & 2,708 & 70 & 3,327 & 60 & 169,343                      & 454& 153,932                      & 153                      \\
\#Edges  & 5,429 & 2,372                     & 4,732 & 1,852                     & 1,166,243                    & 20,274                    & 10,753,238                   & 3,381                    \\
Sparsity & 0.15\%& 48.41\%                   & 0.09\%& 51.44\%                   & 0.01\%& 9.84\%                    & 0.09\%& 14.44\%                  \\
Storage  & 14.9 MB                      & 0.5 MB                    & 47.1 MB                      & 0.9 MB                    & 100.4 MB                     & 0.9MB                     & 435.5 MB                     & 0.6MB                    \\ \Xhline{1.pt}
\end{tabular}}
\end{table*}

\subsection{Hyper-parameter Sensitivity Analysis (\textbf{Q8})}
\label{app_hyper_sec}

In this section, we examine the impact of hyper-parameter $\alpha$ on our proposed method across various downstream tasks. 
The hyper-parameter $\alpha$ adjusts the weight of the centroid discrimination loss during the training process and  
Fig. \ref{fig:hyper} shows the node classification and link prediction performance on the Cora and Reddit datasets.

We select a broad range of values for $\alpha$ to accommodate the gradient variations between the centroid discrimination loss and the cluster loss. 
The performance on Cora stabilizes when $\alpha$ exceeds 1000, while the best results on Reddit are achieved at 10000. As $\alpha$ increases, the model tends to produce a more uniform distribution of centroids, which benefits both node classification and link prediction tasks.

\subsection{Statistics of Condensed Graphs (\textbf{Q6})}
\label{app_statistic_sec}

In Table \ref{stat}, we compare the properties of condensed graphs to those of original graphs. 
Benefiting from self-supervised learning, the condensed graphs generated by our method not only perform better on downstream tasks but also contain fewer nodes and require significantly less storage.
Additionally, these condensed graphs are denser than the original graphs. Given their considerably smaller scale, this increased density enhances message-passing between nodes, benefiting the GNN performance across diverse downstream tasks.

\end{document}